\documentclass[10pt,journal,compsoc]{IEEEtran}

\usepackage[table]{xcolor}
\usepackage{tikz}  
\usepackage{times}
\usepackage{graphicx}
\usepackage{amsmath}
\usepackage{amssymb}
\usepackage{physics}
\usepackage{pifont}
\usepackage{colortbl}
\usepackage{multirow}

\usepackage{subfiles}
\usepackage{pgfplots} 
\usepackage{pgfplotstable} 
\pgfplotsset{compat=newest} 
\usetikzlibrary{plotmarks} 
\usetikzlibrary{colorbrewer} 
\usepackage{booktabs} 
\usepackage{etoolbox,siunitx} 
\robustify\bfseries
\robustify\itshape
\usepgfplotslibrary{statistics}
\usepackage{paralist} 
\usepackage{comment}
\usepackage{epsfig}
\usepackage{graphicx}
\usepackage{subcaption}
\usepackage{hhline}
\usepackage{makecell}
\usepackage[breaklinks=true,letterpaper=true,colorlinks,bookmarks=false]{hyperref}

\newcommand{\OURS}{Vid2CAD}

\newcommand{\myfirstpar}[1]{\textbf{#1.}}
\newcommand{\mypar}[1]{\hspace{-6mm} \textbf{#1.}}

\newcommand{\rott}[1]{\rotatebox[origin=lB]{90}{#1}}

\newcommand{\sdot}[0]{{\cdot}} 

\begin{document}

\title{\OURS: CAD Model Alignment using Multi-View Constraints from Videos}

\author{Kevis-Kokitsi Maninis*,
        Stefan Popov*,
        Matthias Nie{\ss}ner,
        Vittorio Ferrari

\IEEEcompsocitemizethanks{\IEEEcompsocthanksitem K.-K. Maninis, S. Popov, and V. Ferrari are with Google Research. First two authors contributed equally.\protect
\IEEEcompsocthanksitem M. Nie{\ss}ner is with the Technical University of Munich.}}%

\markboth{}  
{}

\IEEEtitleabstractindextext{%
\begin{abstract}
We address the task of aligning CAD models to a video sequence of a complex scene containing multiple objects. 
Our method can process arbitrary videos and fully automatically recover the 9 DoF pose for each object appearing in it, thus aligning them in a common 3D coordinate frame.
The core idea of our method is to integrate neural network predictions from individual frames with a temporally global, multi-view constraint optimization formulation.
This integration process resolves the scale and depth ambiguities in the per-frame predictions, and generally improves the estimate of all pose parameters.
By leveraging multi-view constraints, our method also resolves occlusions and handles objects that are out of view in individual frames, thus reconstructing all objects into a single globally consistent CAD representation of the scene.
In comparison to the state-of-the-art single-frame method Mask2CAD that we build on, we achieve 
substantial improvements on the Scan2CAD dataset (from 11.6\% to 30.7\% class average 
accuracy). The project page is at~\url{http://www.kmaninis.com/vid2cad}.
\end{abstract}

\begin{IEEEkeywords}
CAD model alignment, 3D reconstruction, video understanding.
\end{IEEEkeywords}}

\maketitle

\IEEEdisplaynontitleabstractindextext
\IEEEpeerreviewmaketitle

\section{Introduction}
\label{sec:intro}

\IEEEPARstart{U}{nderstanding} real-world environments using visual data is at the heart of the computer vision community and it is a key requirement for many applications ranging from robotics to AR/VR scenarios.
With the advent of scalable deep learning methods, we have seen significant progress towards these goals with impressive results on 2D images, including image classification~\cite{krizhevsky12nips, simonyan15iclr, he16cvpr}, segmentation~\cite{long15cvpr,chen18pami}, and detection methods~\cite{girshick15iccv,ren15nips,he17iccv}.
In addition, we have seen promising works towards 3D understanding, for example 3D object reconstruction from a single RGB image using learnt data-driven priors~\cite{gkioxari19iccv,popov20eccv}.
However, despite these impressive developments, obtaining full spatial 3D understanding of a whole scene still remains an extremely challenging task.

On one hand many approaches aim to estimate 3D geometry directly from visual data, for instance by predicting mesh geometry~\cite{wang18eccv,gkioxari19iccv,chen20cvpr},
voxel grids~\cite{choy16eccv,girdhar16eccv,wu16nips,xie20ijcv}
or using implicit surface functions~\cite{mescheder19cvpr,park19cvpr}.
On the other hand, another line of research leverages object priors from 3D CAD models datasets~\cite{izadinia17cvpr,huang18eccv,kuo20eccv,tatarchenko19cvpr}.
Their main idea is to formulate image understanding as a joint detection and retrieval problem, where reconstruction relies on nearest neighbor retrieval of 3D models from the dataset. This leads to a simpler, lighter weight model architecture compared to methods directly predicting 3D geometry, and can even provide higher fidelity.

However, this direction often reaches limitations when only considering a single image as it is quite difficult to resolve the ambiguity of an object's depth and scale, and to infer spatial arrangements among objects only with learnt priors from 2D input.
The ambiguity arises because there are many combinations of an object's depth and scale (size) values that lead to the same projection on the image (e.g., large but far away from the camera, or small but near the camera).
In this work, we argue that it is sensible to relax the task and utilize a sequence of RGB images since many computer vision applications are not limited to a single image, but can rather rely on a video stream.
While performing the task of 3D scene understanding on videos instead of single RGB images seems more tractable at a first glance, it raises the question of how to efficiently integrate the per-frame predictions of neural networks.

In this paper, we address the question of how to integrate 3D shape retrievals and alignments from individual frames, e.g. obtained by Mask2CAD~\cite{kuo20eccv}, over a series of video frames in order to produce a globally-consistent 3D representation of the whole scene.
We propose \OURS{}, which leverages multi-view consistency constraints to resolve scale and depth ambiguities.
Our key observation is that the ambiguity can be resolved with constraints on the projections on multiple views, as the object size must remain constant across them.
We feed per-frame object pose predictions into a temporally global non-linear least squares formulation which integrates them across views in order to reconstruct the absolute scale and depth of the retrieved object.
This temporal aggregation process also improves the estimates of other pose parameters such as the object's 3D rotation, and the x,y coordinates of its 3D center.
Finally, by leveraging multi-view constraints our method resolves occlusions and handles objects that are out of view in individual frames, thus reconstructing all objects in the scene into a single globally consistent 3D representation.
In summary, given a video, our method automatically recovers the shape and full 9 DoF pose of each object appearing in it (3D rotation, 3D translation, and scaling along all 3 axes).

We perform extensive experiments on the challenging Scan2CAD dataset~\cite{avetisyan19cvpr}, featuring videos of complex indoor scenes with multiple objects. In comparison to the state-of-the-art single-frame method Mask2CAD that we build on, we achieve a substantial improvement with our temporal integration (from 11.6\% to 30.7\% class average accuracy).
We also compare favorably to a strong alternative we constructed by combining state-of-the-art Multi-View Stereo~\cite{duzceker21deepvideomvs} and RGB-D CAD alignment~\cite{avetisyan19iccv} methods.

\begin{figure*}
\centering
\vspace{-2mm}
\includegraphics[width=.9\linewidth]{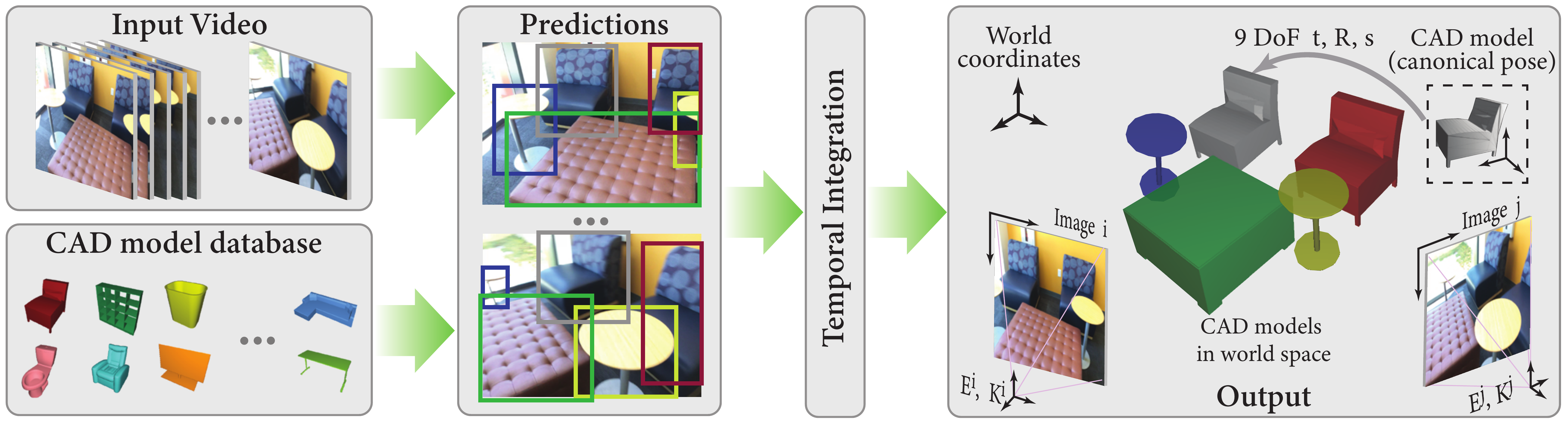} \\[1mm]
\vspace{-2mm}
\caption{{\small
\textbf{Method overview}: given an RGB video sequence, the goal of our method is to find and align a CAD model from a database for each object in the scene. The objective is to find the transformations $t$, $R$, and $s$ that moves each object from its canonical pose to the 3D world coordinate system of the 3D scene. The main idea of our approach is to integrate per-frame neural network predictions with a joint optimization formulation incorporating multi-view constraints. As a result, we obtain a clean, globally-consistent 3D CAD representation of all objects in the scene (right).}}
\label{fig:teaser}
\vspace{-2mm}
\end{figure*}

\section{Related Work}
\label{sec:related}

\myfirstpar{3D from a single image}
Many works in this area~\cite{wang18eccv,mescheder19cvpr,choy16eccv,girdhar16eccv,wu16nips,xie20ijcv,park19cvpr,chen20cvpr}
reconstruct a single object appearing at a fixed 3D position, depth, and scale (i.e., only shape and rotation vary).
Several recent works consider scenes with multiple objects, typically by first detecting them in the 2D image, then reconstructing their 3D shape and pose~\cite{huang18eccv,gkioxari19iccv,kuo20eccv,izadinia17cvpr,tulsiani18cvpr,kundu18cvpr,nie20cvpr,popov20eccv}.
These works compensate for the scale-depth ambiguity in variety of ways, e.g.,
based on estimating an approximate pixelwise depth map from the input image~\cite{huang18eccv},
by requiring manually provided objects' depth and/or scale~\cite{gkioxari19iccv,kuo20eccv} at test time,
or
by estimating the position of a planar floor in the scene and assuming that all objects rest on it~\cite{izadinia17cvpr}.
A few works~\cite{tulsiani18cvpr,nie20cvpr} even attempt to predict object depth and scale directly based on image appearance (which makes them dependent on implicit contextual cues in the overall room appearance).
Finally, CoReNet~\cite{popov20eccv} directly predicts a global 3D scene volume containing all objects in one pass.
However, it has been demonstrated only on scenes with 2 or 3 objects and the monolithic nature of the model makes it unlikely to generalize to more objects.

Our method is mostly related to works based on retrieving the most similar rendering of a CAD model to a 2D detection~\cite{izadinia17cvpr,huang18eccv,kuo20eccv}, out of a given CAD database.
This provides the object's shape and 3D rotation, as well as the x,y coordinates of its center. We propose to resolve for the depth and scale parameters with multi-view integration.

\mypar{3D from multiple views}
Classical works reconstruct a 3D point cloud from multiple views of a scene based on keypoint correspondences~\cite{pollefeys99ijcv,mur15transrobotics,wu133dv,schonberger16cvpr}.
However, the output point cloud is not organized into objects with their semantic labels, 3D shapes, or poses.
Recently FroDO~\cite{runz20cvpr,li2020arxiv} extended this line of works by also detecting objects and reconstructing them in 3D, using both 2D image cues as well as the 3D point cloud.
We tackle the same task, but propose a different multi-view formulation, we directly predict the 9-DoF pose of clean CAD models instead of reconstructing the objects, and have a simpler system that does not require 3D point clouds. Moreover, we show quantitative evaluation on cluttered scenes with multiple objects and multiple classes (ScanNet~\cite{dai17cvpr}, only qualitative with 2 classes in~\cite{runz20cvpr}).
The work of~\cite{qian20eccv} produces volumetric reconstructions of multiple objects in a synthetically generated scene.
ODAM~\cite{li21odam} fits simple super-quadric objects to a video.
Finally,~\cite{ravi20arxiv} reconstructs the shape of a single object given two calibrated views with a neural network.

\mypar{Aligning CAD models using depth and other sensors}
Our work is inspired by techniques for 3D object pose estimation by aligning CAD models to high-quality dense 3D point clouds generated by fusing RGB-D video frames acquired with an additional depth sensor.
Early works use known pre-scanned objects~\cite{salas13cvpr}, hand-crafted features~\cite{nan2012tog,frome04eccv,li2015database,shao2012tog}, and human intervention~\cite{shao2012tog}.
Recent works use deep networks to directly align shapes on the dense point clouds~\cite{avetisyan19cvpr,avetisyan19iccv,avetisyan20eccv,izadinia20cvpr}.

SLAM++~\cite{salas13cvpr} is one of the first works to reconstruct a scene as a set of previously known object shapes. It processes depth maps and aligns objects to them in 6 DoF, while also localizing the position of the cameras. Its optimization objective is based on matching the depth profile of an object surface to the observed metric depth maps of the video frames.

All of the above methods have access to much more and cleaner information than we do, but are limited to videos acquired by an RGB-D sensor.
Requiring only RGB videos opens up the possibility of operating on a much larger pool of videos, e.g. from YouTube.
The task then becomes more challenging, since without the depth sensor, the Z-depth position and the 3 DoF anisotropic scaling of the objects must be estimated via multi-view cues.
In Sec.~\ref{sec:exp:mvsrgbd}, we compare to a hybrid method constructed by replacing the clean 
RGB-D point-clouds with reconstructions by Multi-View Stereo~\cite{duzceker21deepvideomvs} in 
the state-of-the-art CAD alignment method~\cite{avetisyan19iccv}.
 
Fei et al.~\cite{fei18eccv} align a known set of shapes on a video in 4 DoF, by using a camera with an inertial sensor. As with a depth sensor, this reduces the search space for alignment. In our work we solve for 9 DoF alignment.

\section{Method}
\label{sec:method}

\begin{figure}
\centering
\vspace{-2mm}
\includegraphics[width=1\linewidth]{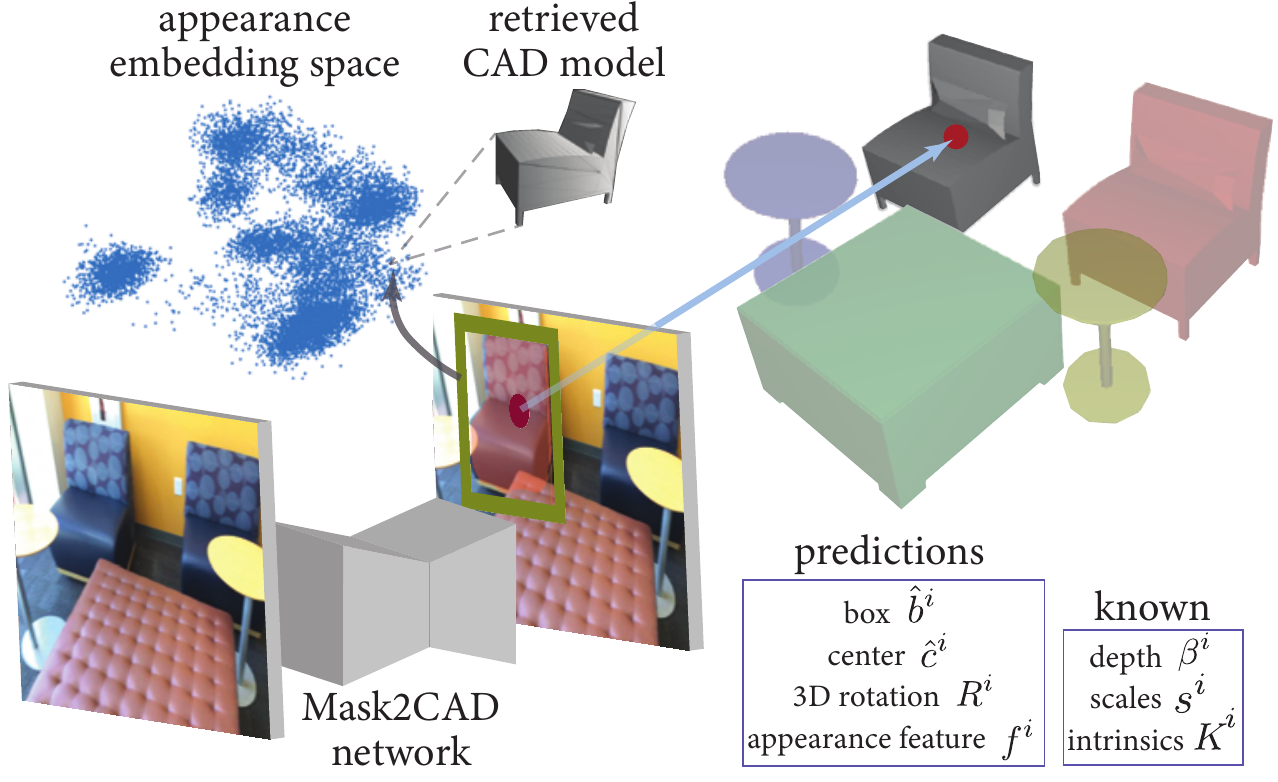} \\[1mm]
\vspace{-4mm}
\caption{{\small
\textbf{The Mask2CAD method:} On top of the traditional 2D instance segmentation outputs (box, class, mask), Mask2CAD predicts the 2D projection $\hat{c}^i$ of the 3D object center on the image, the 3D rotation matrix $R^i$, and the shape code vector $f^i$. However, it requires the depth  $\beta^i$ of the center and the scaling transformation $s^i$ as input.}}
\label{fig:mask2cad}
\end{figure}

\begin{table}
    \centering
    {\small
    \begin{tabular}{rl}
    \hline
    $t$ & $3\times 1$ translation CAD $\rightarrow$ world \\
    $R$ & $3\times 3$ rotation CAD $\rightarrow$ world  \\
    $s$ & $3\times 1$ scaling CAD $\rightarrow$ world \\
    $i$ & frame index\\
    $R^i$ & $3\times 3$ rotation CAD $\rightarrow$ camera view space \\
    $s^i$ & $3\times 1$ scaling CAD $\rightarrow$ world  \\
    $E^i$ & $3\times 4$ extrinsic camera matrix \\
    $ E^i_{R}, e^i_{t}$ & $3\times 3$ rotation and $3\times 1$ translation of $E^i$.\\
    $K^i$ & $3\times 3$ intrinsic camera matrix \\
    $\hat{c}^i$ & $2\times 1$ object center in the image  \\
    $\hat{b}^i$ & $2\times 1$ amodal box in the image \\
    $\beta^i$ & $1\times 1$ depth value of the object center \\
    \hline\\
    \end{tabular}
    }
    \vspace{-5mm}
    \caption{{\small
    \label{tbl:symbols}
     Math notation. The first three rows determine the 9 DoF object pose we want to reconstruct. We use the superscript $^i$ for entities attached to frame $i$, and $\hat{~}$ for 2D vectors on an image.}}
    \vspace{-2mm}
\end{table}

Our goal is to align CAD models from a database to a video of a scene (Fig.~\ref{fig:teaser}).
For each object, we want to find which CAD model corresponds to it, and a 9 Degrees-of-Freedom (DoF) transformation that maps from its initial database pose to the 3D world (scene).
We seek for a 3 DoF rotation matrix $R$, a 3 DoF translation vector $t$, and a 3 DoF anisotropic scaling vector {\small$s=(s_x, s_y, s_z)^T$} such that the vertices $v$ of the CAD model are placed in their correct position in the world by applying the transformation:
{\small
\begin{align}
h(v) &= t + R\cdot s\cdot v\label{eq::object_to_world}
\end{align}
}
The CAD models live in a canonical space in the database (scale-normalized to a constant size, centered at the origin, and in a canonical orientation common to all objects within a class).
The object projects to image $i$ by {\small $\hat{v}^i=K^i\cdot (e^i_{t} +E^i_{R}\cdot h(v))$}.
We assume that we know the pose of the camera w.r.t the world at each video frame $i$ (extrinsic calibration matrix {\small $E^i=\left[ E^i_{R} | e^i_{t}\right]$}) as well as the projection function to the image (intrinsic calibration matrix $K^i$).
Extrinsic parameters can be obtained from off-the-shelf SfM methods such as~\cite{schonberger16cvpr, mur15transrobotics}. In our evaluation, we use the provided extrinsics, consistent with the most recent methods~\cite{runz20cvpr, li2020arxiv}.
Tab.~\ref{tbl:symbols} summarizes our notation.

In Sec.~\ref{sec:mask2cad}, we first review how the task can be (partially) addressed given a single image by the state-of-the-art method Mask2CAD~\cite{kuo20eccv}. We discuss its shortcomings and then propose a solution that leverages multi-view constraints induced by the video (Sec.~\ref{sec:temporal_integration_core}).
In Sec.~\ref{sec:method:scale-from-single-frame}, we discuss an extension involving predicting approximate object scales from a single frame based on recognition.

\subsection{Base method: Mask2CAD}
\label{sec:mask2cad}

\myfirstpar{The technique}
Mask2CAD~\cite{kuo20eccv} is based on a semantic instance segmentation model~\cite{li17cvpr, he17iccv}, which detects objects of a predefined set of classes in an image $i$.
For each detection, Mask2CAD predicts 2D properties (i.e., 2D bounding box, class, confidence score, and segmentation mask), as well as some 3D properties: rotation $R^i$, the 2D projection $\hat{c}^i \in \mathbb{R}^2$
of the 3D center on the image, and a shape code vector $f^i$.
The latter is used to compare natural images of objects to synthetic images of the CAD models.
During inference Mask2CAD retrieves the most similar CAD model from the database based on similarity in an appearance embedding space. This CAD model is then placed in the world by using the predicted rotation $R^i$, while translating the object by moving the predicted center $\hat{c}^i$ to a manually-given depth value $\beta^i$ by using the intrinsic matrix $K^i$ (Fig.~\ref{fig:mask2cad}). The size of the object is also manually provided through the known vector $s^i$.

\mypar{Strengths and limitations}
As Mask2CAD casts 3D object reconstruction as retrieval of clean CAD models, it naturally outputs high-quality shapes, without the need to address over-smoothing or tessellation artifacts typical of methods that predict 3D geometry directly from the image (e.g., voxel grids~\cite{choy16eccv,girdhar16eccv,wu16nips,xie20ijcv}, meshes~\cite{wang18eccv,gkioxari19iccv,chen20cvpr} or point clouds~\cite{fan17cvpr,mandikal18bmvc}).

\begin{figure}
\centering
\vspace{-2mm}
\includegraphics[width=.95\linewidth]{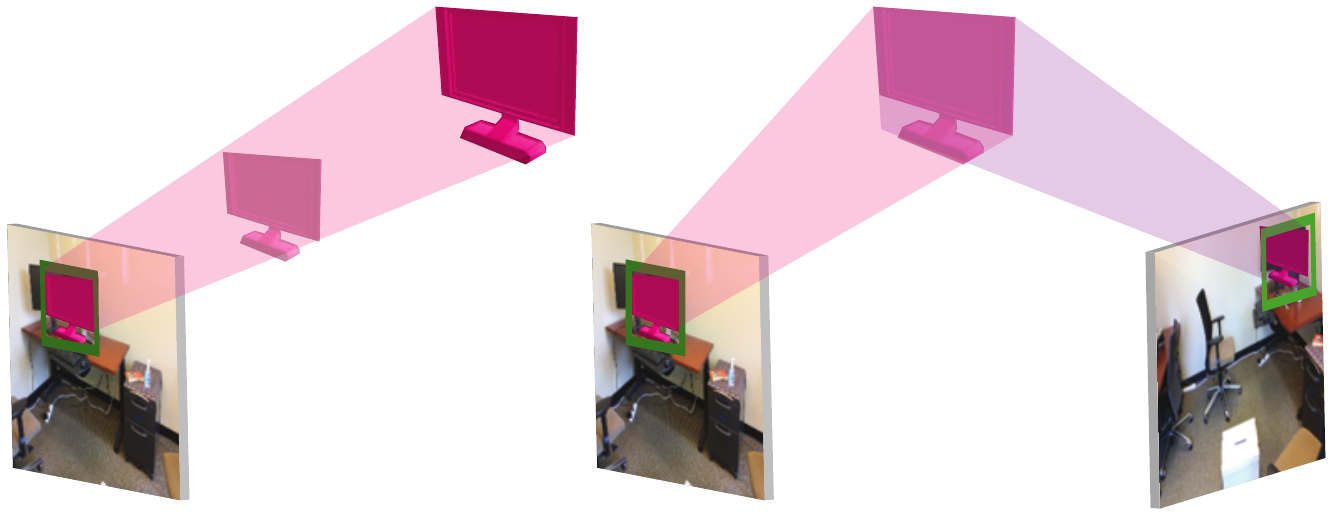} \\[1mm]
\vspace{-4mm}
\caption{{\small 
\textbf{Scale-depth ambiguity:}
(Left) Placing a small object near the camera or a larger copy of the same object far from it lead to the same projection on the image.
(Right) We address this by leveraging multi-view constraints.}}
\label{fig:scale_depth_ambiguity}
\vspace{-3mm}
\end{figure}

However, given a single image, Mask2CAD is not able to infer the size of the objects nor their position along the $z$ axis (depth $\beta$), due to the scale-depth ambiguity.
The ambiguity arises because of the projection from 3D to 2D (Fig.~\ref{fig:scale_depth_ambiguity}, left).
By simultaneously changing the size of an object and its position along the depth axis, we can obtain the same projection on the image (e.g., a small object near the camera, or a large object far from it).
This scale-depth ambiguity is an inherent limitation for 3D reconstruction methods from a single image, which need to compensate for it in various ways (Sec.~\ref{sec:related}).

In practice, Mask2CAD as well as Mesh-RCNN~\cite{gkioxari19iccv} work around this limitation by using the ground-truth depth  $\beta^i$ and the size of the objects during inference (as the database-to-world scaling transformation $s^i$).
In real settings, this information is not available at test time and such methods are not usable 
automatically.

\subsection{Temporal integration}
\label{sec:temporal_integration_core}

\begin{figure*}
\centering
\vspace{-2mm}
\includegraphics[width=.95\linewidth]{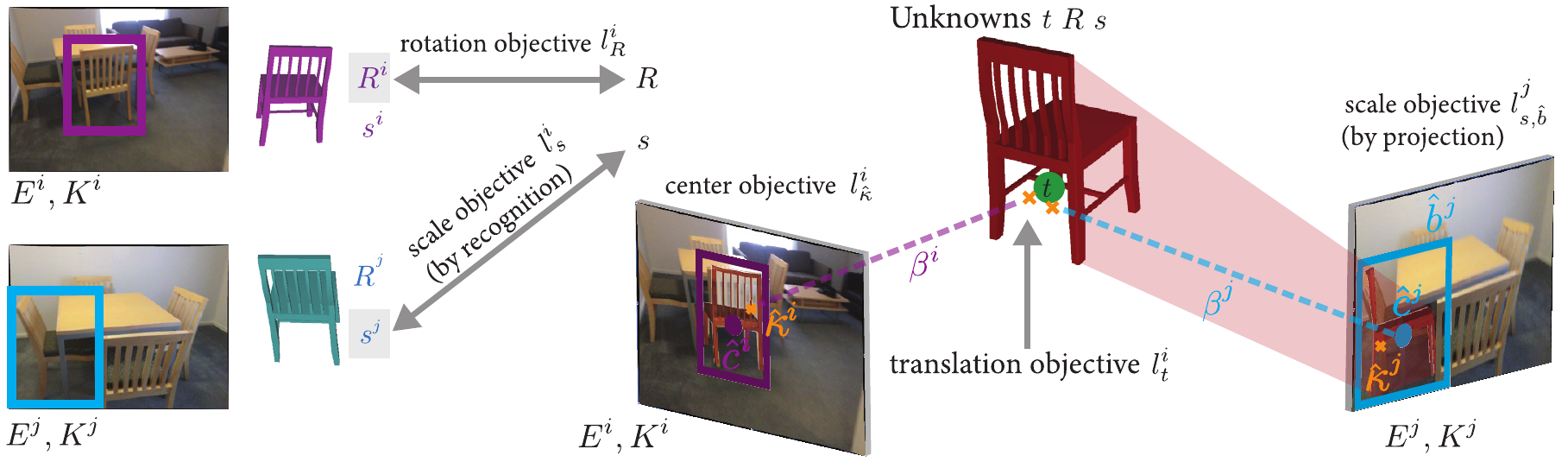} \\[1mm]
\vspace{-4mm}
\caption{{\small\textbf{Temporal Integration:}
We formulate our task as a constrained optimization problem with objectives that arise from multi-view constraints, given by the input frames (left).
The center objective~\eqref{eq::center_objective} keeps the value of the auxiliary variable $\hat{\kappa}^i$ close to the predicted box center $\hat{c}^i$ in frame $i$ (center of figure).
The translation objective~\eqref{eq::translation_objective} maintains the consistency between the desired 3D object center $t$ and the center $(\hat{\kappa}^i_x, \hat{\kappa}^i_y, \beta^i)$ formed by the auxiliary variables and the desired depth $\beta^i$.
The rotation objective~\eqref{eq::rotation_objective} relates the desired rotation $R$ to the rotations $R^i$ predicted in each frame (top-left image).
Finally, the scale objectives~\eqref{eq::scale_from_boxes},~\eqref{eq::scale_from_recognition} constrain the desired scaling transformation $s$ based on the predicted box ($\hat{b}^j$ in the right image) and the predicted scalings $s^i$ (bottom-left image), respectively.}
}
\label{fig:temporal_integration}
\vspace{-2mm}
\end{figure*}

We propose to integrate the single-frame Mask2CAD predictions across frames in a video, as they offer multiple views of the same objects.
This integration process brings several advantages:
(1) it resolves the scale-depth ambiguity, inferring both of them automatically (Fig.~\ref{fig:scale_depth_ambiguity}, right);
(2) it improves the estimates of other pose parameters such as the object’s 3D rotation, and the x,y coordinates of its 3D center;
(3) it resolves occlusions and objects that are out of view in individual frames, allowing
to create one globally consistent 3D reconstruction of the scene (rather than a separate partial reconstruction for each frame).

Different video frames offer different views of the same scene and thus different
constraints on the translation $t$, rotation $R$, and scaling $s$ transformations of an object.
We formulate temporal integration as an optimization problem, applied to one object at a time.
For each video frame $i$, our method inputs the Mask2CAD predictions for that object in frame $i$, i.e., rotation $R^i$, 2D projection $\hat{c}^i$ of center, shape code $f^i$, and 2D bounding box $\hat{b}^i$.
We output a single integrated CAD model selection and a full 9 DoF pose $(t,R,s)$ mapping it to the world.
Below we explain how.

\mypar{Selecting a CAD model}
Each frame votes with its predicted shape code vector $f^i$, with weight proportional to the object detection score of Mask2CAD in that frame.
We select the CAD model with the highest vote.

\mypar{Optimization formulation}
We use hard constraints as well as soft-constraint terms arising from relaxing multi-view geometric constraints imposed by relating a single desired output pose to multiple predictions from the individual frames (which are naturally noisy).
In the following we present the terms~\eqref{eq::center_objective},~\eqref{eq::translation_objective},~\eqref{eq::rotation_objective},~\eqref{eq::scale_from_boxes} separately first, and then combine them into our overall optimization objective~\eqref{eq:overall-objective}.

\mypar{Constraints for translation $t$}
The 3D object center must project near the 2D centers $\hat{c}^i$ predicted by
Mask2CAD in each frame, thus inducing multi-view constraints for the object translation $t$ (Fig.~\ref{fig:temporal_integration}).
All CAD models are pre-processed, so that they are centered at the origin of their canonical space. Applying~\eqref{eq::object_to_world} for objects at the origin, instead of a variable object center, removes the dependency of the center on rotation $R$ and scaling $s$.
Hence, the object center in world space becomes equal to $t$:
{\small
\begin{align}
t + R \sdot s \sdot (0, 0, 0)^T = t + (0, 0, 0)^T = t
\end{align}
}
To create the constraints, we model the object center as seen from each frame $i$
using 3 auxiliary variables -- the 2D position in image space
$\hat{\kappa}^i=(\hat{\kappa}^i_x, \hat{\kappa}^i_y)$ and the depth $\beta^i$ with
respect to frame $i$ (Fig.~\ref{fig:temporal_integration}, right).
We add a soft-constraint that keeps $\hat{\kappa}^i$ close to the 2D center $\hat{c}^i$ predicted by Mask2CAD.
We transform $(\hat{\kappa}^i_x, \hat{\kappa}^i_y, \beta^i)$ to world space and
add another soft constraint that keeps the resulting constructed 3D center
close to the desired object center $t$ (which we are looking for).
Therefore, the 2D center objective $l_{\hat{\kappa}}^i$
and the 3D translation objective $l_t^i$ are:
{\small
\begin{align}
l_{\hat{\kappa}}^i &= \norm{\hat{\kappa}^i - \hat{c}^i}_{L_1}\label{eq::center_objective} \\
l_t^i &= \norm{
  (E^i_R)^{-1} (K^i)^{-1} \sdot (
    \beta^i\hat{\kappa}^i_x, \beta^i\hat{\kappa}^i_y, \beta^i)^T
  -e^i_t
  - t
}_{L_1}\label{eq::translation_objective}
\end{align}
}
In our overall objective, we will minimize~\eqref{eq::center_objective} over $\hat{\kappa}^i$, while minimizing~\eqref{eq::translation_objective} over $\beta^i$, $\hat{\kappa}^i$, and $t$. Thus, we need multiple frames to avoid degenerate solutions for $t$.

Modeling centers per-frame relaxes the projection equations and improves reconstruction performance compared to projecting the 3D object center $t$ to all frames and comparing to $\hat{c}^i$ directly.
In the latter case, frames where the object is close to the camera get a large weight due to the division by a small depth value during projection.
We also add a hard constraint keeping the centers above a minimum depth in each frame: {\small $\beta^i > 0.1 \textrm{m}$}.

\mypar{Constraints for rotation $R$}
To create constraints for $R$, we note that there are two ways to transform the object from database space to the 3D coordinate system of the camera in a frame $i$ (camera view space):
(1) move the object into world space through the 9 DoF pose transformation $(t,R,s)$ and then into camera view space through the extrinsic parameters $E^i$; or
(2) directly use the rotation matrix $R^i$ predicted by Mask2CAD from frame $i$ and combine it with the translation vector $t^i$ from database space to camera view space.
Both ways lead to the same result:
{\small
\begin{align}
e^i_{t} + E^i_{R}\sdot(t + R \sdot s \sdot v ) = t^i + R^i \sdot s \sdot v
\end{align}
}
This equation is valid for any point on the object. Assuming non-degenerate
transformations, this can only be true if $E^i_{R} \sdot R = R^i$.
We use this to create a soft constraint, the rotation objective $l_{R}^i$ for each frame:
{\small
\begin{align}
l_{R}^i = \norm{R^i - E^i_{R} \sdot R}_{L_2}\label{eq::rotation_objective}
\end{align}
}
We ensure $R$ remains a valid rotation matrix during the optimization process by adding a hard 
constraint keeping its corresponding quaternion normalized.

For vertically symmetric objects, we look for any valid rotation by considering only the minimum distance of the predicted rotation to all valid rotations in the objective~\eqref{eq::rotation_objective}.

\mypar{Constraints for scaling $s$}
To infer the anisotropic scaling transformation $s$, and thus the size of the objects in 3D, we use the 2D amodal bounding boxes $\hat{b}^i$ predicted by Mask2CAD (Fig.~\ref{fig:temporal_integration}).
Since the scaling affects the projection of the CAD model vertices on the image, we design constraints so that $s$ leads to projections respecting these boxes.
Specifically, for a candidate value of $s$, we first project the vertices $v$ of the CAD model on frame $i$ based on $s$:
{\small
\begin{align}
\hat{v}^i = K^i\cdot \left(e^i_t + E^i_R \sdot (t + R \sdot s \sdot v)\right)
\end{align}
}
We apply this transformation to all vertices and compute the bounding box $\hat{v}^i_{box}$ around the resulting 2D points $\hat{v}^i$.
Then, we soft-constrain this box to match $\hat{b}^i$, resulting in the objective:
{\small
\begin{align}
    l^i_{s,\hat{b}} &= \mathrm{d_{box}} (\hat{v}^i_{box}, \hat{b}^i)
    \label{eq::scale_from_boxes}
\end{align}}
where $\mathrm{d_{box}}$ is the $L_1$ distance between the box sides (left, right, top, bottom).
In addition to the unknown $s$, this objective also depends on other unknowns $R,t$. During 
optimization, we jointly solve for all unknowns simultaneously.

\mypar{Overall optimization objective}
The full objective $l^i$ for frame $i$ is formulated as a weighted sum of the objectives above, and the total sum over all frames is:
{\small
\begin{align}
    L = \sum_i l^i = \sum_i a_tl_t^i +  a_{\hat{\kappa}}l_{\hat{\kappa}}^i + a_{R}l_{R}^i + a_{s,\hat{b}}l^i_{s,\hat{b}}\label{eq:overall-objective}
\end{align}} 
We jointly minimize the objective $L$ over the desired 9 DoF transformation
($R,t,s$), as well as over the auxiliary variables $\hat{\kappa}^i$ and $\beta^i$ that we introduced.
The optimization is subject to the two hard constraints we formulated above (i.e., {\small $\beta^i>0.1$m} and the rotation quaternion normalization).
The objective function has L1 and L2 terms in the variables being optimized, which we optimize using 
gradient descent (initialized with {\small $t=(0, 0, 0), s=(1, 1, 1)$}, identity rotation {\small $R$, 
$\hat{\kappa}^i$} to the center of the image, and {\small $\beta=1$m}).
We set the hyper-parameter weights $a$ as described in Sec.~\ref{sec:exp}.

\subsection{Predicting object scale from a single frame}
\label{sec:method:scale-from-single-frame}

\myfirstpar{Scale from recognition.}
Due to the scale-depth ambiguity one cannot determine the 3D scale and depth of an {\em arbitrary} object from a single image. However, if the object class is known, one can use the average class size as a rough estimate~\cite{zhang20eccv}.
We can go a step further by noticing that the size of an object depends on its particular model within a class, which can be estimated based on its 2D appearance alone, i.e., by recognition.
We exploit this by augmenting Mask2CAD with a head to directly predict the scaling factor $s$ mapping the CAD model to the world, for each detected box. For better results, we use a separate scale regressor specialized for each class.
Note that vanilla Mask2CAD already predicts a class for each box, which we use to select which regressor output to take.

\mypar{Using single-frame scale in temporal integration}
Predicting object scales by recognition can benefit temporal integration.
We add to~\eqref{eq:overall-objective} a term encouraging the output object scaling $s$ to be close to the scalings $s^i$ predicted in the individual frames:
{\small
\begin{align}
l_s^i = \norm{s - s^i}_{L_1}\label{eq::scale_from_recognition}
\end{align}
}
Note that inferring scalings within our temporal integration method based on projection on amodal 2D boxes~\eqref{eq::scale_from_boxes} or based on recognition~\eqref{eq::scale_from_recognition}
are complementary and work best when used together (Sec.~\ref{sec:exp:full-system}).

\mypar{Deriving object depth from a single image}
By having a prediction $s$ for the size of the object from a single frame, we can also infer the depth of the object by minimizing the reprojection error of the CAD model on the predicted amodal 2D bounding box (analog to~\eqref{eq::scale_from_boxes}, but this time optimized over depth). A related technique was also presented in~\cite{kundu18cvpr}.
While in theory this trick addresses the scale-depth ambiguity even from a single frame, in practice the estimated depth values are quite unstable, as they are strongly affected by small inaccuracies in the predicted amodal box, object scale, predicted rotation, and/or predicted object center.
As we show in Sec.~\ref{sec:exp}, quantitative results are much better when using temporal integration.

\subsection{Implementation details}
\label{sec:implementation-details}

\myfirstpar{Mask2CAD architecture}
We use the default settings for the network architecture, which builds on the ShapeMask~\cite{kuo19iccv} instance segmentation method with ResNet-50~\cite{he16cvpr} backbone.
For the added scale prediction branch we used 4 convolution blocks with a fully-connected output layer that outputs $3\cdot N_{cls}$ outputs for the class-specific anisotropic scalings (with $N_{cls}$ the number of classes).

\mypar{Temporal association}
Our temporal integration method (Sec.~\ref{sec:temporal_integration_core}) inputs the predictions of Mask2CAD for one object across multiple frames.
As Mask2CAD detects objects independently in each frame,
we first automatically associate detections of one physical object across frames using a standard 
tracking-by-detection approach (Sec. 4.2 in~\cite{marin14ijcv}).
However, some objects go out of view and re-appear later on, causing fragmented tracks.
We fix this issue by clustering in 3D space the object alignments produced by our temporal integration method from the initial tracks.
We form the first cluster by picking the object with the highest detection score and adding all 
objects of the same class within a fixed distance to it ($40$cm translation, $40^\circ$ rotation, and 
$40\%$ scale).
We repeat this process, forming more clusters until no object remains.

After clustering, we re-run our temporal integration on each cluster, this time using all information in 
all tracks within it. This improves the estimated 9DOF pose of the objects as this second temporal 
integration sees more views of the same object at once.

\section{Experiments}
\label{sec:exp}

\begin{table*}[th]
	\small
	\centering
	\fontsize{9}{11}
	\selectfont
	\begin{tabular}{c|c|cccc||>{\columncolor[gray]{0.9}}c>{\columncolor[gray]{0.9}}c|ccccccccccc}
		\textbf{\rott{Family}}  &\textbf{id} & \rott{depth} & \rott{scales} & \rott{association} & \rott{rot. 
		sym.} & \rott{\textbf{class avg.}} & \rott{\textbf{global avg.}} & \rott{bathtub}& \rott{bookshelf}& 
		\rott{cabinet}& \rott{chair}& \rott{display}& \rott{sofa}& \rott{table}& \rott{trashbin}& \rott{other} 
		\\
		\hline
		\multirow{5}{*}{\rott{\begin{tabular}[c]{@{}c@{}}Single-frame\\baselines\end{tabular}}} 
		&$b_1$   & gt   & gt      & gt  & -  & 33.7 & 41.2 & 24.2 & 25.0 & 24.2 & 68.9 & 48.2 & 31.9 & 
		28.9 & 48.7 & 3.4 \\
		&$b_2$   & avg  & avg     & gt  & -  & 2.5  & 3.8  & 0.0  & 1.9  & 1.5  & 7.7  & 4.7  & 1.8  & 1.4  & 
		2.6  & 1.2 \\
		&$b_3$   & avg  & avg      & thr & -  & 2.5  & 3.5  & 0.0  & 1.9  & 1.5  & 6.8  & 3.7  & 2.7  & 1.4  & 
		3.0  & 1.2 \\
		&$b_4$   & deriv  & pred   & gt  & -  & 12.1 & 16.9 & 9.2  & 2.8  & 6.9  & 33.2 & 17.3 & 6.2  & 
		7.2  
		& 25.4 & 0.5 \\
		&$b_5$   & deriv & pred    & thr & -  & 11.6 & 16.0 & 8.3  & 3.8  & 5.4  & 30.9 & 17.3 & 5.3  & 
		7.1  
		& 25.9 & 0.5 \\
		\hline
		\multirow{5}{*}{\rott{\begin{tabular}[c]{@{}c@{}}Temporal\\Integration\end{tabular}}}& 
		$\mathbf{F}$     & mv & mv+pred    & track & yes  & 30.7 & 38.6 & 28.3 & 12.3 & 23.8 & 64.6 & 
		37.7 
		& 26.5 & 28.9 & 47.8 & 6.6 \\\hhline{|~|----------------|}
		&$a_1$   & mv & mv      & gt  & no   & 33.5 & 41.0 & 25.8 & 20.8 & 30.4 & 65.6 & 26.7 & 33.6 & 
		35.1 & 60.8 & 2.7 \\
		&$a_2$   & mv & pred    & gt  & no   & 30.2 & 38.0 & 25.0 & 15.6 & 20.4 & 65.8 & 40.8 & 19.5 
		& 
		23.3 & 58.6 & 2.4 \\
		&$a_3$   & mv & mv+pred & gt  & no   & 37.4 & 44.9 & 38.3 & 17.9 & 33.5 & 73.3 & 39.8 & 
		32.7 
		& 31.5 & 63.4 & 6.1 \\
		&$a_4$   & mv & mv+pred & gt  & yes  & 37.6 & 45.2 & 38.3 & 17.9 & 33.5 & 73.3 & 39.8 & 
		32.7 & 32.2 & 64.7 & 6.1 \\
		\hline
		MVS + RGB-D &$M$   & MVS & pred & - & yes  & 18.8 & 21.7 & 15.8 & 8.5 & 17.3 & 34.3 & 25.7 
		& 
		15.0 & 10.9 & 35.8 & 6.1 \\
	\end{tabular}
	
	\caption{ \small Quantitative evaluation on the Scan2CAD dataset~\cite{avetisyan19cvpr}.
		We compare our multi-view integration methods ($F$ and $a$) to single-frame baselines ($b$) 
		and and the MVS + RGB-D method ($M$).
		Method $b_5$ is the best fully automated single-frame baseline, and $F$ is our fully automated 
		temporal integration method.
		The shortcuts are: ground-truth (gt), average (avg), predicted (pred), derived based on scale 
		and reprojection (deriv), estimated based on multi-view constraints (mv).
		See main body text for details.
	}
	\label{tbl:exp}
	
\end{table*}

\myfirstpar{Datasets and evaluation metric}
We use videos from ScanNet~\cite{dai17cvpr}, 3D CAD models from ShapeNetCore~\cite{chang15arxiv}, and annotations connecting them from Scan2CAD~\cite{avetisyan19cvpr}.
ScanNet provides RGB-D videos of rich indoor scenes with multiple objects in complex spatial arrangements.
It also provides camera parameters for individual frames and dense depth fusion~\cite{niessner2013real,dai2017bundlefusion} reconstructions. {\em We only use the RGB videos and the camera parameters, ignoring all depth data}.
ShapeNetCore provides CAD models for 55 object classes, in a canonical orientation within a class.
Scan2CAD provides manual 9 DoF alignments of ShapeNetCore models onto ScanNet scenes for $9$ super-classes.

We use these data sets both for training and for evaluation.
During training, we consider all ScanNet videos in the official train split whose scenes have Scan2CAD annotations (1194 videos). For training the Mask2CAD network, we take individual video frames and we project the aligned CAD models onto them.
We set the weights $a$ of the optimization objective~\eqref{eq:overall-objective} empirically on the 
same training set by using grid search, resulting in weights: $a_t=20$, $a_{\hat{\kappa}}=3 $, 
$a_{R}=0.1 $, $a_{s,\hat{b}}=3$. These weights are kept fixed in all experiments for all videos.

We evaluate our method and the baselines on the 312 videos in ScanNet's val split, containing 3184 objects.
We quantify performance using the Scan2CAD evaluation protocol~\cite{avetisyan19cvpr}: a 
ground-truth 3D object is considered accurately detected if one of the model outputs matches its 
class and 9 DoF alignment (satisfying all error thresholds \emph{at the same time}: $20\%$ scale, 
$20^\circ$ rotation, and $20$cm translation).
We report accuracy averaged over classes (`class avg.') as well as over all object instances (`global avg').

\mypar{Training Mask2CAD}
We train Mask2CAD for 96000 iterations with the same data augmentations as in~\cite{kuo20eccv} (HSV-color, ROI, and image scale jittering). The initial learning rate is set to 0.8 and is reduced by a factor of 10 at $2/3$ and $5/6$ of the total number of iterations.
We include objects that are partially visible and whose center is truncated during training, as it improves performance.

\subsection{Single-frame baselines - Mask2CAD}
\label{sec:exp:single-frame}

\myfirstpar{Original Mask2CAD}
We evaluate several variants of our single-frame Mask2CAD baseline.
The first one ($b_1$) is the original setting from~\cite{kuo20eccv}, using ground-truth depth and object size at test time to tackle the scale-depth ambiguity.
It also relies on the ground-truth 2D object boxes at test time: for each ground-truth box it only 
keeps the most overlapping detected box (if it overlaps $>0.3$). All other detections are discarded. 
We call this procedure `ground-truth association'.
This model is directly given 4 out of the 9 DoF as ground-truth at test time (1 depth and 3 scales). It also benefits from the cleanup made by ground-truth association, which indirectly provides some information about 2 other DoFs (x,y coordinates of the center).
Having access to so much ground-truth at test time is unrealistic. In the following we explore several variants of Mask2CAD which use less of it.

\mypar{More automatic variants}
As variant $b_2$, we estimate an object depth and scale by taking the average scale and depth of its class from the training set. This does not require altering Mask2CAD's architecture.

An arguably better way to estimate scale and depth automatically is our idea from Sec.~\ref{sec:method:scale-from-single-frame}: we extend MaskCAD's architecture to predict object scale and then use it to derive depth by reprojection on the 2D box (variant $b_4$).

For both ways to get scale and depth we consider using ground-truth association or not. In the latter case detections are simply filtered at $0.2$ score, which leads to fully automatic models $b_3$, $b_5$.

\mypar{Duplicate removal} 
Detections for the same physical object in different frames result in multiple copies in the output, which might lower the performance metrics for these single-frame baselines. Hence, we use our 3D clustering algorithm from Sec.~\ref{sec:implementation-details} to remove such duplicate detections (i.e., keeping only the top-scored one in each cluster).
Note that all single-frame baselines process \emph{all} frames of the video and produce a single 3D reconstruction for the entire scene, containing all objects detected in all frames together.

\mypar{Results (Tab.~\ref{tbl:exp})}
Model $b_1$ achieves $33.7\%$ accuracy, which can be seen as a theoretical upper bound of Mask2CAD as it uses substantial ground-truth information at test time.
Using class-average depth and scale values instead of ground-truth leads to poor results, reaching only $2.5\%$ accuracy for models $b_2,b_3$. This is not surprising, as the evaluation metric demands rather accurate poses.
Our extension from Sec.\ref{sec:method:scale-from-single-frame} allows to predict object depth and scale by recognition, improving performance substantially to $12.1\%$ ($b_4$) and $11.6\%$ ($b_5$). The difference due to using ground-truth association is small ($0.5\%$ from $b_4$ to $b_5$).
Model $b_5$ represents the best fully automatic variant of Mask2CAD we built, and is the reference model to improve further upon with temporal integration.

\subsection{Temporal integration}
\label{sec:exp:full-system}

\begin{figure*}
\centering

\includegraphics[width=.98\linewidth]{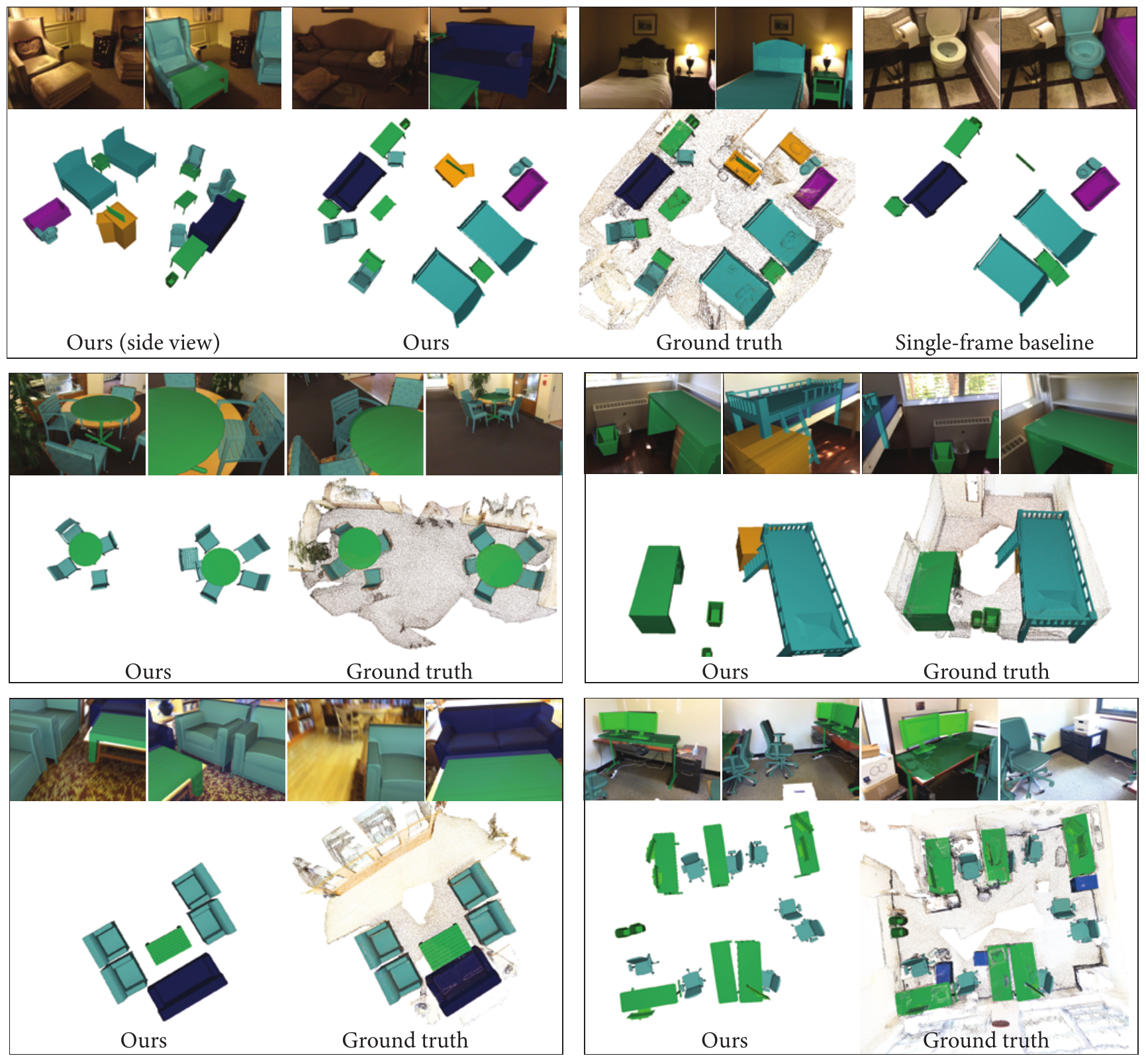} \\[1mm]

\caption{\small \textbf{Qualitative results:}
We compare the alignment produced by our temporal integration method to the ground-truth and to the best automatic single-frame baseline (top); i.e., our extended Mask2CAD, cf. Tab.~\ref{tbl:exp}, $b_5$.
We also show our alignments overlaid on the input frames, which highlight the difficulty of the 
problem as only a small part of the scene is visible in each frame.
}
\label{fig:qualitative}
\end{figure*}

\myfirstpar{Our fully-automated method}
Our full method $F$ uses all objective function terms in~\eqref{eq:overall-objective} and~\eqref{eq::scale_from_recognition}, and performs temporal association with a tracker (Sec.~\ref{sec:implementation-details}). It is fully automatic as it does not use any ground-truth at test time.
It achieves $30.7\%$ accuracy, $2.6\times$ better ($+19.1\%$) than the best automatic single-frame method $b_5$ (which already included our enhancements from Sec.~\ref{sec:method:scale-from-single-frame}).
These comparisons demonstrate the dramatic improvements brought by our main contribution.
We show qualitative results in Fig.~\ref{fig:qualitative}.

\mypar{Ablation study}
We study the effect of varying the way we estimate object scale during temporal integration, and the use of a rotation symmetry term. This results in 4 settings ($a_1$-$a_4$) and we show their performance in Tab.~\ref{tbl:exp}.

The first way ($a_1$) is to estimate scale based only on multi-view reprojection constraints~\eqref{eq::scale_from_boxes} on 2D detection boxes.
The second way ($a_2$) is to only use the single-frame Mask2CAD scale predictions via the term~\eqref{eq::scale_from_recognition}.
The first way performs better by $+3.2\%$, highlighting the power of multi-view constraints.
Moreover, using both ways ($a_3$) improves accuracy further by $+3.9\%$, showing that the two mechanism are complementary.
Finally, taking into account vertically symmetric objects ($a_4$) as described in Sec.~\ref{sec:temporal_integration_core} improves only by a small amount ($+0.2\%$).

For this ablation we used ground-truth association instead of the automatic tracker. As in Sec.~\ref{sec:exp:single-frame}, this matches detections to ground-truth boxes and it brings perfect temporal association (via the 3D object id associated to a box in the annotations of Scan2CAD).
This keeps the study focused on the differences brought by the various terms in our core method, removing secondary effects due to the tracker and our track merging mechanism (Sec.~\ref{sec:implementation-details}).
It also allows to estimate the margin for improvement when using better tracking algorithms:
$a_4$ is only moderately better than our fully-automatic method $F$ ($+6.9\%$).
Finally, $a_4$ can be fairly compared to single-frame baseline $b_4$, as they both use the same ground-truth association. Our temporal integration brings massive improvements also in this case ($+25.5\%$).

\begin{figure}
\begin{center}
  \includegraphics[width=0.99\linewidth]{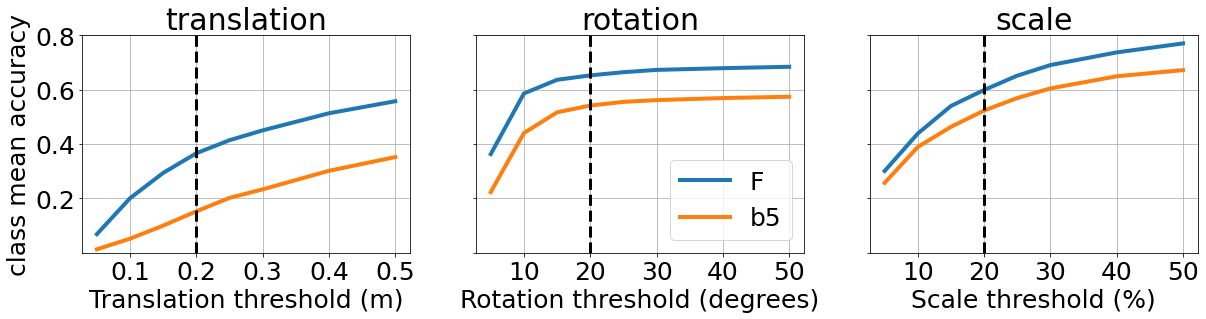}
  \includegraphics[width=0.99\linewidth]{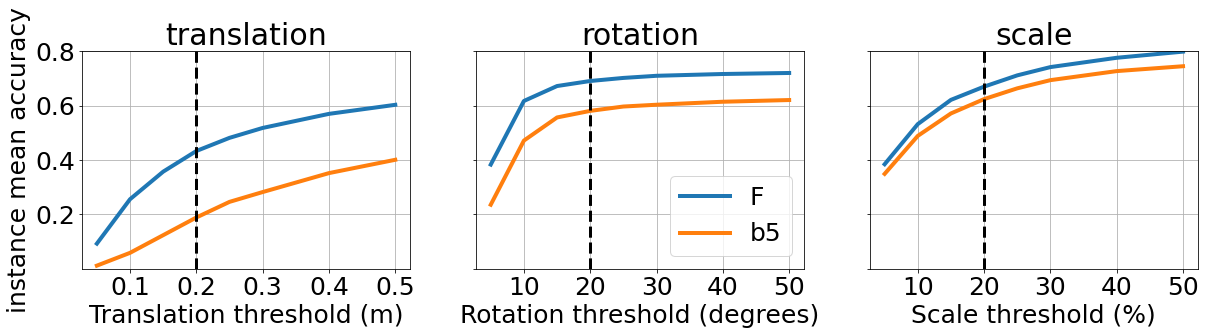}
\end{center}
  \caption{\small Class mean accuracy and instance mean accuracy as a function of the evaluation 
  threshold, for our fully automatic method ($F$) as well as the best-performing baseline ($b_5$). 
  We examine each transformation type separately. The dotted line indicates the default error 
  threshold.}
\label{fig:eval_metric}
\end{figure}

\mypar{Performance for each transformation type}
In Fig.~\ref{fig:eval_metric} we report the performance for each type of transformation separately and swipe the error threshold. We compare our best automatic method $F$ to the best-performing automatic baseline $b_5$. We report class average accuracy, and the vertical dotted line indicates the default error threshold. The main bottleneck for accuracy is the translation error, which is expected since our system does not use depth as input, while rotation and scale are predicted more accurately. Translation is also the transformation where our fully automatic method improves the most over $b_5$, which proves the effectiveness of our multi-view formulation.

\mypar{Computational cost}
Our temporal integration optimization formulation is very lightweight, it naturally parallelizes over objects, and operates on only 40 frames uniformly spaced over the video.
Hence, it does not take significant runtime (2.5s total \textit{per video}) compared to running Mask2CAD on every frame (0.2s \textit{per frame}).

Thanks to its speed, Vid2CAD with temporal integration can be operated online as well.
We can run multi-view optimization repeatedly, e.g. once every 5 seconds, on the portion of the 
video seen so far. Every time we can update the estimated object poses, and even include new 
objects that recently appeared.

\mypar{Accuracy of CAD model retrieval}
To isolate the accuracy of CAD model retrieval, we evaluate only on the objects that satisfy the 
alignment error thresholds. We compute the IoU of a retrieved CAD model with respect to the 
ground truth object, both placed in the canonical pose. The baseline $b5$ achieves mIoU of 73.1\%, 
whereas our method $F$ achieves 85\%, showing that integrating information from multiple frames 
also helps to retrieve a more accurate CAD model.

\subsection{Comparison to MVS + RGB-D CAD alignment}
\label{sec:exp:mvsrgbd}
Modern methods for aligning CAD models to video use a dedicated RGB-D depth sensor to acquire a high-quality dense 3D point-cloud of the scene via depth-fusion. Thanks to this, they can directly align CAD models on the point-cloud~\cite{avetisyan19cvpr, avetisyan19iccv, avetisyan20eccv, izadinia20cvpr}.
Instead, our method only uses the RGB frames.

In this experiment we explore how well the best RGB-D CAD alignment method~\cite{avetisyan19iccv} would perform without the benefits of a depth sensor, by replacing its input with point-clouds reconstructed by the most recent state-of-the-art Multi-View Stereo method DVMVS~\cite{duzceker21deepvideomvs}.
We train DVMVS on ScanNet, and re-train~\cite{avetisyan19iccv} on its output.
This effectively enables~\cite{avetisyan19iccv} to operate on purely RGB videos at test time, constructing a strong alternative to our method.
For both~\cite{duzceker21deepvideomvs,avetisyan19iccv}, we obtained the code and training guidelines from the authors.

As shown in Tab.~\ref{tbl:exp} (method $M$), this delivers 18.8\% class average accuracy on the Scan2CAD val set, clearly below our full method $F$ (30.7\%).
After careful visual inspection, we found most failure cases to occur on objects whose surfaces are inaccurately reconstructed by DVMVS.
In contrast, our method works directly on the video frames and bypasses MVS entirely.

We also note that the pipeline of DVMVS +~\cite{avetisyan19iccv} requires stronger supervision than our method.
DVMVS needs ground-truth depth for training (here on Scan2CAD itself).
Moreover,~\cite{avetisyan19iccv} needs the alignments of the CAD models \emph{on the 3D scene} 
for training. Instead,
our method can train directly from CAD alignments \emph{on the 2D image}, which are easier to 
annotate (as done for Pix3D~\cite{sun18pix3d}).

In a summary, this experiment demonstrates that aligning CAD models to RGB video is a truly challenging task that cannot be solved simply by applying existing RGB-D alignment methods on top of off-the-shelf MVS stereo (even when both ingredients are state-of-the-art).
Our method offers a different kind of solution, which performs substantially better.

For transparency, we also compare to~\cite{avetisyan19cvpr,avetisyan19iccv} in their original form, i.e. inputting clean RGB-D scans.
Surprisingly, our fully automatic method $F$ performs on par with~\cite{avetisyan19cvpr} ($35.6\%$ class avg, $31.7\%$ global avg; vs our $30.7\%/38.6\%$, Tab.~\ref{tbl:exp}).
However, the state-of-the-art RGB-D CAD alignment~\cite{avetisyan19iccv} performs even better ($44.6\%/50.7\%$).

\subsection{Comparison to ODAM~\cite{li21odam}}
The concurrent work \cite{li21odam} proposed to populate the scene with posed super-quadrics.
Different to our work, they fit simpler super-quadric shapes instead of full CAD models, and their alignments are in 7 DoF (rotation is predicted only around the ``up'' axis).

We compare to~\cite{li21odam} with their detection-based metrics using their implementation: precision, recall, and F1 score at a pre-defined Intersection-over-Union (IoU) threshold.
A 3D bounding box is considered a true positive if the Intersection-over-Union (IoU) between itself and a ground-truth box in the same object class is above a predefined threshold.

The results are presented in Table~\ref{tab:odam}.
Vid2CAD outperforms ODAM in the stricter IoU thresholds ($IoU > 0.5$ and $IoU > 0.7$), which 
highlights the effectiveness of our multi-view optimization.

\begin{table}
    \centering
    \footnotesize
    \setlength{\tabcolsep}{7pt}
    \begin{tabular}{l|ccc|}
               & \multicolumn{3}{c}{Precision/Recall/F1 @ IoU} \\
        Method & IoU$>0.25$  & IoU$>0.5$ & IoU$>0.7$  \\
        \hline
        ODAM~\cite{li21odam} & \textbf{64.7}/\textbf{58.6}/\textbf{61.5}   & 31.2/28.3/29.7 & 3.8/3.5/3.6\\ 
        Vid2CAD (ours) & 56.9/55.7/56.3 & \textbf{34.2}/\textbf{33.5}/\textbf{33.9} & \textbf{10.7}/\textbf{10.4}/\textbf{10.5} \\
    \end{tabular}

    \caption{Quantitative results on ScanNet using the ODAM metric. Vid2CAD outperforms ODAM as 
    the IoU threshold gets stricter, providing more accurate results.}
    \label{tab:odam}

\end{table}

\section{Conclusions}
\label{sec:conclusions}

We introduced \OURS, a method to align CAD models to a video of a complex scene containing multiple objects.
Our core idea is to integrate per-frame network predictions across time by leveraging multi-view constraints, thus obtaining a globally-consistent CAD representation of the 3D scene.
Compared to the best single-frame method Mask2CAD, we achieve a substantial improvement, from 11.6\% to 30.7\% class average accuracy.
Future work includes joint camera pose estimation and CAD alignment, as well as supporting 
dynamic environments.

\mypar{Acknowledgements}
We thank Weicheng Kuo for providing us with detailed information about Mask2CAD, and for helping 
us to train and evaluate it, Kejie Li for helping us with the evaluation of ODAM, and  
\mbox{Angela~Dai} for her contribution in the supplemental 
video. This work was supported by the ERC Starting Grant Scan2CAD (804724).


\ifCLASSOPTIONcaptionsoff
  \newpage
\fi

\bibliographystyle{ieee.bst}
\bibliography{references}

\begin{thebibliography}{10}\itemsep=-1pt

\bibitem{avetisyan19cvpr}
A.~Avetisyan, M.~Dahnert, A.~Dai, M.~Savva, A.~X. Chang, and M.~Nie{\ss}ner.
\newblock {Scan2CAD}: Learning cad model alignment in {RGB-D} scans.
\newblock In {\em CVPR}, 2019.

\bibitem{avetisyan19iccv}
A.~Avetisyan, A.~Dai, and M.~Nie{\ss}ner.
\newblock End-to-end cad model retrieval and 9dof alignment in 3d scans.
\newblock In {\em ICCV}, 2019.

\bibitem{avetisyan20eccv}
A.~Avetisyan, T.~Khanova, C.~Choy, D.~Dash, A.~Dai, and M.~Nie{\ss}ner.
\newblock {SceneCAD}: Predicting object alignments and layouts in {RGB-D}
  scans.
\newblock In {\em ECCV}, 2020.

\bibitem{chang15arxiv}
A.~X. Chang, T.~Funkhouser, L.~Guibas, P.~Hanrahan, Q.~Huang, Z.~Li,
  S.~Savarese, M.~Savva, S.~Song, H.~Su, et~al.
\newblock Shapenet: An information-rich 3d model repository.
\newblock {\em arXiv:1512.03012}, 2015.

\bibitem{chen18pami}
L.-C. Chen, G.~Papandreou, I.~Kokkinos, K.~Murphy, and A.~Yuille.
\newblock {DeepLab}: Semantic image segmentation with deep convolutional nets,
  atrous convolution, and fully connected {CRFs}.
\newblock {\em TPAMI}, 2018.

\bibitem{chen20cvpr}
Z.~Chen, A.~Tagliasacchi, and H.~Zhang.
\newblock Bsp-net: Generating compact meshes via binary space partitioning.
\newblock In {\em CVPR}, 2020.

\bibitem{choy16eccv}
C.~B. Choy, D.~Xu, J.~Gwak, K.~Chen, and S.~Savarese.
\newblock {3D-R2N2}: A unified approach for single and multi-view {3D} object
  reconstruction.
\newblock In {\em ECCV}, 2016.

\bibitem{dai17cvpr}
A.~Dai, A.~X. Chang, M.~Savva, M.~Halber, T.~Funkhouser, and M.~Nie{\ss}ner.
\newblock Scannet: Richly-annotated 3d reconstructions of indoor scenes.
\newblock In {\em CVPR}, 2017.

\bibitem{dai2017bundlefusion}
A.~Dai, M.~Nie{\ss}ner, M.~Zollh{\"o}fer, S.~Izadi, and C.~Theobalt.
\newblock Bundlefusion: Real-time globally consistent 3d reconstruction using
  on-the-fly surface reintegration.
\newblock {\em ACM Transactions on Graphics (ToG)}, 36(4):1, 2017.

\bibitem{duzceker21deepvideomvs}
A.~Duzceker, S.~Galliani, C.~Vogel, P.~Speciale, M.~Dusmanu, and M.~Pollefeys.
\newblock Deepvideomvs: Multi-view stereo on video with recurrent
  spatio-temporal fusion.
\newblock In {\em CVPR}, 2021.

\bibitem{fan17cvpr}
H.~Fan, H.~Su, and L.~J. Guibas.
\newblock A point set generation network for 3d object reconstruction from a
  single image.
\newblock In {\em CVPR}, 2017.

\bibitem{fei18eccv}
X.~Fei and S.~Soatto.
\newblock Visual-inertial object detection and mapping.
\newblock In {\em ECCV}, 2018.

\bibitem{frome04eccv}
A.~Frome, D.~Huber, R.~Kolluri, T.~B{\"u}low, and J.~Malik.
\newblock Recognizing objects in range data using regional point descriptors.
\newblock In {\em ECCV}, 2004.

\bibitem{girdhar16eccv}
R.~Girdhar, D.~Fouhey, M.~Rodriguez, and A.~Gupta.
\newblock Learning a predictable and generative vector representation for
  objects.
\newblock In {\em ECCV}, 2016.

\bibitem{girshick15iccv}
R.~Girshick.
\newblock Fast {R-CNN}.
\newblock In {\em ICCV}, 2015.

\bibitem{gkioxari19iccv}
G.~Gkioxari, J.~Malik, and J.~Johnson.
\newblock Mesh {R-CNN}.
\newblock In {\em ICCV}, 2019.

\bibitem{he17iccv}
K.~He, G.~Gkioxari, P.~Doll{\'a}r, and R.~Girshick.
\newblock Mask {R-CNN}.
\newblock In {\em ICCV}, 2017.

\bibitem{he16cvpr}
K.~He, X.~Zhang, S.~Ren, and J.~Sun.
\newblock Deep residual learning for image recognition.
\newblock In {\em CVPR}, 2016.

\bibitem{huang18eccv}
S.~Huang, S.~Qi, Y.~Zhu, Y.~Xiao, Y.~Xu, and S.-C. Zhu.
\newblock Holistic {3D} scene parsing and reconstruction from a single {RGB}
  image.
\newblock In {\em ECCV}, 2018.

\bibitem{izadinia20cvpr}
H.~Izadinia and S.~M. Seitz.
\newblock Scene recomposition by learning-based icp.
\newblock In {\em CVPR}, 2020.

\bibitem{izadinia17cvpr}
H.~Izadinia, Q.~Shan, and S.~M. Seitz.
\newblock {Im2CAD}.
\newblock In {\em CVPR}, 2017.

\bibitem{krizhevsky12nips}
A.~Krizhevsky, I.~Sutskever, and G.~E. Hinton.
\newblock Imagenet classification with deep convolutional neural networks.
\newblock In {\em NIPS}, 2012.

\bibitem{kundu18cvpr}
A.~Kundu, Y.~Li, and J.~M. Rehg.
\newblock {3D-RCNN}: Instance-level 3d object reconstruction via
  render-and-compare.
\newblock In {\em CVPR}, 2018.

\bibitem{kuo20eccv}
W.~Kuo, A.~Angelova, T.-Y. Lin, and A.~Dai.
\newblock {Mask2CAD}: {3D} shape prediction by learning to segment and
  retrieve.
\newblock In {\em ECCV}, 2020.

\bibitem{kuo19iccv}
W.~Kuo, A.~Angelova, J.~Malik, and T.-Y. Lin.
\newblock Shapemask: Learning to segment novel objects by refining shape
  priors.
\newblock In {\em ICCV}, 2019.

\bibitem{li21odam}
K.~Li, D.~DeTone, Y.~F.~S. Chen, M.~Vo, I.~Reid, H.~Rezatofighi, C.~Sweeney,
  J.~Straub, and R.~Newcombe.
\newblock Odam: Object detection, association, and mapping using posed rgb
  video.
\newblock In {\em ICCV}, 2021.

\bibitem{li2020arxiv}
K.~Li, H.~Rezatofighi, and I.~Reid.
\newblock Mo-ltr: Multiple object localization, tracking, and reconstruction
  from monocular {RGB} videos.
\newblock {\em arXiv:2012.05360}, 2020.

\bibitem{li2015database}
Y.~Li, A.~Dai, L.~Guibas, and M.~Nie{\ss}ner.
\newblock Database-assisted object retrieval for real-time 3d reconstruction.
\newblock In {\em Computer Graphics Forum}, volume~34. Wiley Online Library,
  2015.

\bibitem{li17cvpr}
Y.~Li, H.~Qi, J.~Dai, X.~Ji, and Y.~Wei.
\newblock Fully convolutional instance-aware semantic segmentation.
\newblock In {\em CVPR}, 2017.

\bibitem{long15cvpr}
J.~Long, E.~Shelhamer, and T.~Darrell.
\newblock Fully convolutional models for semantic segmentation.
\newblock In {\em CVPR}, 2015.

\bibitem{mandikal18bmvc}
P.~Mandikal, N.~K. L., M.~Agarwal, and V.~B. Radhakrishnan.
\newblock 3d-lmnet: Latent embedding matching for accurate and diverse 3d point
  cloud reconstruction from a single image.
\newblock In {\em BMVC}, 2018.

\bibitem{marin14ijcv}
M.~J. Marin-Jimenez, A.~Zisserman, M.~Eichner, and V.~Ferrari.
\newblock Detecting people looking at each other in videos.
\newblock {\em IJCV}, 106(3):282--296, 2014.

\bibitem{mescheder19cvpr}
L.~Mescheder, M.~Oechsle, M.~Niemeyer, S.~Nowozin, and A.~Geiger.
\newblock Occupancy networks: Learning 3d reconstruction in function space.
\newblock In {\em CVPR}, 2019.

\bibitem{mur15transrobotics}
R.~Mur-Artal, J.~M.~M. Montiel, and J.~D. Tardos.
\newblock {ORB-SLAM}: a versatile and accurate monocular slam system.
\newblock {\em IEEE transactions on robotics}, 2015.

\bibitem{nan2012tog}
L.~Nan, K.~Xie, and A.~Sharf.
\newblock A search-classify approach for cluttered indoor scene understanding.
\newblock {\em ACM Transactions on Graphics (TOG)}, 2012.

\bibitem{nie20cvpr}
Y.~Nie, X.~Han, S.~Guo, Y.~Zheng, J.~Chang, and J.~J. Zhang.
\newblock Total3dunderstanding: Joint layout, object pose and mesh
  reconstruction for indoor scenes from a single image.
\newblock In {\em CVPR}, 2020.

\bibitem{niessner2013real}
M.~Nie{\ss}ner, M.~Zollh{\"o}fer, S.~Izadi, and M.~Stamminger.
\newblock Real-time 3d reconstruction at scale using voxel hashing.
\newblock {\em ACM Transactions on Graphics (ToG)}, 32(6):1--11, 2013.

\bibitem{park19cvpr}
J.~J. Park, P.~Florence, J.~Straub, R.~Newcombe, and S.~Lovegrove.
\newblock Deepsdf: Learning continuous signed distance functions for shape
  representation.
\newblock In {\em CVPR}, 2019.

\bibitem{pollefeys99ijcv}
M.~Pollefeys, R.~Koch, and L.~Van~Gool.
\newblock Self-calibration and metric reconstruction inspite of varying and
  unknown intrinsic camera parameters.
\newblock {\em IJCV}, 32(1):7--25, 1999.

\bibitem{popov20eccv}
S.~Popov, P.~Bauszat, and V.~Ferrari.
\newblock {CoReNet}: Coherent {3D} scene reconstruction from a single {RGB}
  image.
\newblock In {\em ECCV}, 2020.

\bibitem{qian20eccv}
S.~Qian, L.~Jin, and D.~F. Fouhey.
\newblock Associative3d: Volumetric reconstruction from sparse views.
\newblock In {\em ECCV}, 2020.

\bibitem{ravi20arxiv}
N.~Ravi, J.~Reizenstein, D.~Novotny, T.~Gordon, W.-Y. Lo, J.~Johnson, and
  G.~Gkioxari.
\newblock Accelerating 3d deep learning with pytorch3d.
\newblock {\em arXiv preprint arXiv:2007.08501}, 2020.

\bibitem{ren15nips}
S.~Ren, K.~He, R.~Girshick, and J.~Sun.
\newblock Faster {R-CNN}: Towards real-time object detection with region
  proposal networks.
\newblock In {\em NIPS}, 2015.

\bibitem{runz20cvpr}
M.~Runz, K.~Li, M.~Tang, L.~Ma, C.~Kong, T.~Schmidt, I.~Reid, L.~Agapito,
  J.~Straub, S.~Lovegrove, et~al.
\newblock Frodo: From detections to 3d objects.
\newblock In {\em CVPR}, 2020.

\bibitem{salas13cvpr}
R.~F. Salas-Moreno, R.~A. Newcombe, H.~Strasdat, P.~H. Kelly, and A.~J.
  Davison.
\newblock {SLAM}++: Simultaneous localisation and mapping at the level of
  objects.
\newblock In {\em CVPR}, 2013.

\bibitem{schonberger16cvpr}
J.~L. Sch\"{o}nberger and J.-M. Frahm.
\newblock Structure-from-motion revisited.
\newblock In {\em CVPR}, 2016.

\bibitem{shao2012tog}
T.~Shao, W.~Xu, K.~Zhou, J.~Wang, D.~Li, and B.~Guo.
\newblock An interactive approach to semantic modeling of indoor scenes with an
  {RGBD} camera.
\newblock {\em ACM Transactions on Graphics (TOG)}, 2012.

\bibitem{simonyan15iclr}
K.~Simonyan and A.~Zisserman.
\newblock Very deep convolutional networks for large-scale image recognition.
\newblock In {\em ICLR}, 2015.

\bibitem{sun18pix3d}
X.~Sun, J.~Wu, X.~Zhang, Z.~Zhang, C.~Zhang, T.~Xue, J.~B. Tenenbaum, and W.~T.
  Freeman.
\newblock Pix3d: Dataset and methods for single-image 3d shape modeling.
\newblock In {\em CVPR}, 2018.

\bibitem{tatarchenko19cvpr}
M.~Tatarchenko, S.~R. Richter, R.~Ranftl, Z.~Li, V.~Koltun, and T.~Brox.
\newblock What do single-view 3d reconstruction networks learn?
\newblock In {\em CVPR}, 2019.

\bibitem{tulsiani18cvpr}
S.~Tulsiani, S.~Gupta, D.~Fouhey, A.~A. Efros, and J.~Malik.
\newblock Factoring shape, pose, and layout from the 2d image of a 3d scene.
\newblock In {\em CVPR}, 2018.

\bibitem{wang18eccv}
N.~Wang, Y.~Zhang, Z.~Li, Y.~Fu, W.~Liu, and Y.-G. Jiang.
\newblock Pixel2mesh: Generating 3d mesh models from single {RGB} images.
\newblock In {\em ECCV}, 2018.

\bibitem{wu133dv}
C.~Wu.
\newblock Towards linear-time incremental structure from motion.
\newblock In {\em 3DV}, 2013.

\bibitem{wu16nips}
J.~Wu, C.~Zhang, T.~Xue, W.~T. Freeman, and J.~B. Tenenbaum.
\newblock Learning a probabilistic latent space of object shapes via {3D}
  generative-adversarial modeling.
\newblock In {\em NIPS}, 2016.

\bibitem{xie20ijcv}
H.~Xie, H.~Yao, S.~Zhang, S.~Zhou, and W.~Sun.
\newblock Pix2vox++: multi-scale context-aware 3d object reconstruction from
  single and multiple images.
\newblock {\em IJCV}, 128(12):2919--2935, 2020.

\bibitem{zhang20eccv}
J.~Y. Zhang, S.~Pepose, H.~Joo, D.~Ramanan, J.~Malik, and A.~Kanazawa.
\newblock Perceiving 3d human-object spatial arrangements from a single image
  in the wild.
\newblock In {\em ECCV}, 2020.

\end{thebibliography}

\end{document}